\documentclass{article}

\usepackage{arxiv}

\usepackage[utf8]{inputenc}
\usepackage[T1]{fontenc}
\usepackage{hyperref}
\usepackage{url}
\usepackage{booktabs}
\usepackage{amsfonts}
\usepackage{amsmath}
\usepackage{amssymb}
\usepackage{nicefrac}
\usepackage{microtype}
\usepackage{cleveref}
\usepackage{graphicx}
\usepackage{natbib}
\usepackage{doi}
\usepackage{multirow}
\usepackage{tabularx}
\usepackage{subcaption}
\usepackage{placeins}

\title{Is a 9B Decision Foundation Model Needed for Symptom-to-Disease Classification? Fine-Tuning Clef-Flash Against TF-IDF Baselines}

\author{
  \textbf{S. M. Mahbub Uddin Ahmed} \\
  Department of Computer Science and Engineering \\
  United International University \\
  \texttt{sahmed221068@bscse.uiu.ac.bd}
}

\renewcommand{\shorttitle}{Fine-Tuning Clef-Flash Against TF-IDF Baselines}
\renewcommand{\headeright}{}
\renewcommand{\undertitle}{}
\date{}

\hypersetup{
  pdftitle={Is a 9B Decision Foundation Model Needed for Symptom-to-Disease Classification? Fine-Tuning Clef-Flash Against TF-IDF Baselines},
  pdfsubject={cs.CL, cs.AI, q-bio.QM},
  pdfauthor={S. M. Mahbub Uddin Ahmed},
  pdfkeywords={Clinical NLP, Decision Foundation Models, Clef-Flash, Parameter-Efficient Fine-Tuning, QLoRA, Explainable AI, Counterfactual Perturbation, Feature Attribution},
}

\begin{document}
\maketitle

\begin{abstract}
Automated disease classification from patient-reported symptom descriptions is commonly approached either with lightweight discriminative models, which are fast but rely on surface lexical features, or with autoregressive large language models (LLMs), which offer broad language understanding but generate text token by token, adding latency and requiring output parsing. Drawing loosely on the dual-process (System~1 / System~2) view of clinical reasoning, we study decision foundation models, which score a closed set of candidate labels in a single forward pass without generating text. We adapt Clef-Flash, a 9-billion parameter decision model built on a Qwen3.5 backbone, to the 24-class \textit{Symptom2Disease} benchmark using 4-bit quantized low-rank adaptation (QLoRA) with differential learning rates on a stratified split of 960 training and 240 test cases. The fine-tuned model reaches \textbf{95.42\% accuracy} (95\% CI 91.98--97.42\%) and a \textbf{macro F1 score of 0.9551}, up from 65.00\% for the same unadapted weights run locally (62.92\% through the remote API). However, TF-IDF linear SVM and logistic regression baselines trained on the same split reach 98.75\% (95\% CI 96.39--99.57\%) with sub-millisecond CPU inference, so on this benchmark the fine-tuned model offers no accuracy advantage over a well-tuned linear classifier. To probe what the model relies on, we apply a Leave-One-Symptom-Out (LOSO) clause-ablation analysis to four illustrative narratives. In a psoriasis example, removing the clause describing silvery scales lowers the predicted probability from above 99.99\% to 0.12\% and shifts the prediction to impetigo (99.59\%), whereas removing any other clause changes it by less than 0.001 percentage points. These results show that decision architectures can deliver structured, single-pass predictions from large language backbones, but that short, lexically distinctive benchmarks such as \textit{Symptom2Disease} cannot demonstrate an advantage over simple baselines; validation on longer, real clinical text and clinician review are needed.
\end{abstract}


\section{Introduction}
\label{sec:intro}

\begin{figure*}[t]
\centering
\includegraphics[width=\textwidth]{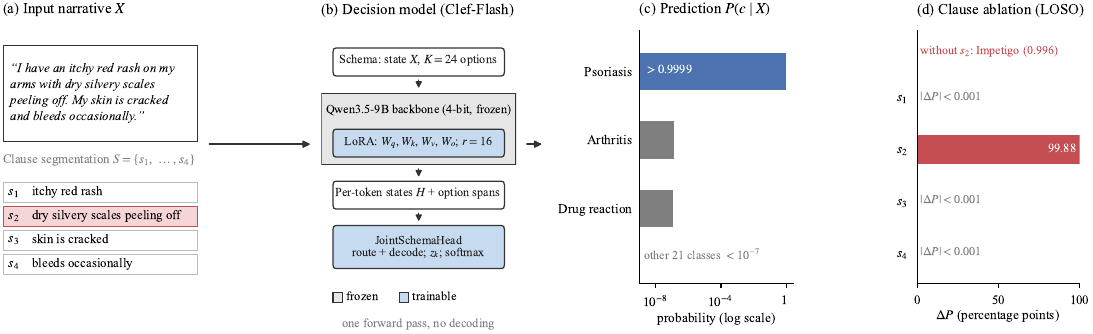}
\caption{\textbf{Overview of the decision pipeline and clause-level attribution.} 
(a)~A first-person symptom narrative is used as raw text input and segmented into symptom clauses $s_1,\dots,s_4$ for attribution.
(b)~Clef-Flash: a 4-bit frozen Qwen3.5-9B backbone with trainable LoRA adapters on the attention projections, and the trainable JointSchemaHead, which routes per-token evidence to option representations and scores all $K=24$ candidate conditions in one non-autoregressive forward pass, without generating tokens.
(c)~Predicted distribution for this narrative (log scale): Psoriasis receives a probability above $0.9999$.
(d)~Leave-One-Symptom-Out ablation: removing $s_2$, \textit{``dry silvery scales peeling off''}, lowers the Psoriasis probability by 99.88 percentage points and flips the prediction to Impetigo (99.59\%), whereas removing any other clause changes it by less than 0.001 points. All values are measured model outputs for this one illustrative, author-written narrative.}
\label{fig:overview}
\end{figure*}

Distilling free-text patient descriptions into an ordered differential diagnosis is an important initial step during patient intake, primary care triage, and telemedicine consultations \citep{singhal2023large, hasan2026clinllm}. While computational decision support systems have long been developed to assist clinical staff and reduce diagnostic oversights, existing solutions tend to fall into two distinct groups, each with clear practical drawbacks.

\paragraph{Dual-Process Cognition and Clinical Decision Support.}
Human diagnostic reasoning is often described through dual-process cognitive theory \citep{croskerry2009clinical, kahneman2011thinking}. Under this view, clinical judgment combines \textit{System 1}, fast and largely intuitive pattern recognition, with \textit{System 2}, slower analytical reasoning used in more deliberate differential workups. We use this distinction only as a loose analogy. Many LLM-based approaches to diagnosis frame the task generatively, producing explanations token by token. This offers linguistic flexibility but adds latency (one such pipeline reports up to 2.5 seconds per case \citep{hasan2026clinllm}), can produce unsupported statements, and requires downstream parsing of free-text output. For high-volume intake and routing, a fast, structured ranking of candidate diagnoses over a closed label set, together with an inspectable account of which reported symptoms drove it (\Cref{fig:overview}), may be a useful complement to, not a replacement for, deliberative clinical reasoning.

Traditional discriminative models, including Support Vector Machines (SVMs) \citep{cortes1995support}, Random Forests \citep{breiman2001random}, and shallow recurrent neural networks \citep{ijacsa2025disease, sbmufdsdd2026}, frame disease prediction as closed-set multi-class classification. These methods are fast and produce deterministic output scores. On short texts with distinctive vocabulary they can be highly accurate, as our own baselines confirm (\Cref{tab:scorecard}). However, their bag-of-words or task-specific representations carry no prior language knowledge, which is expected to limit them on paraphrase, negation, and longer or noisier clinical text.

In contrast, pretrained generative language models such as GPT-4 \citep{achiam2023gpt4}, LLaMA-3 \citep{dubey2024llama3}, and Med-PaLM \citep{singhal2023large} possess broad medical knowledge and flexible syntactic comprehension. Nevertheless, relying on autoregressive token generation for clinical decision making introduces substantial operational hurdles. Autoregressive decoders can generate ungrounded claims, incorrect assertions, or misplaced diagnostic certainty. Furthermore, clinical intake pipelines require structured outputs, such as diagnostic codes and class probabilities, yet generative models frequently introduce conversational preambles or malformed serialization that precipitate downstream parsing failures. Generating text token by token also requires sequential forward passes; the CLIN-LLM pipeline, for example, reports processing each case in under 2.5 seconds on an NVIDIA A100 GPU \citep{hasan2026clinllm}, which can become a bottleneck in high-volume triage settings. Finally, attributing token-level evidence within autoregressive decoders remains difficult, complicating clinical verification and regulatory auditing \citep{doshivelez2017rigorous}.

Decision foundation models offer an alternative approach to this trade-off, as demonstrated by Cloudflare's Clef and Clef-Flash architectures \citep{cloudflare2026clef}. Rather than employing an autoregressive text decoder, Clef-Flash pairs a 9-billion parameter transformer backbone (Qwen3.5-9B) \citep{qwen35model} with a small transformer scoring head, the \textbf{JointSchemaHead} \citep{cloudflare2026clef}. The model evaluates all candidate disease hypotheses in a single non-autoregressive forward pass, producing normalized class probabilities without generating intermediate text tokens (\Cref{fig:overview}). The released backbone includes a vision encoder, but our investigation focuses on unstructured natural language narratives.

In this paper, we adapt Clef-Flash to the 24-disease \textit{Symptom2Disease} benchmark \citep{barman2023symptom2disease} and analyze its decision behavior. Our main contributions are:
\begin{itemize}
    \item \textbf{Parameter-Efficient Clinical Adaptation}: We fine-tune Clef-Flash using 4-bit Quantized Low-Rank Adaptation (QLoRA) on the attention layers alongside differential learning rates for the JointSchemaHead. On a stratified 80/20 split (240 test cases), the model achieves \textbf{95.42\% accuracy} (95\% CI 91.98--97.42\%; macro F1 0.9551). This is a 30-point gain over the unadapted model.
    \item \textbf{Same-Split Classical Baselines}: We train TF-IDF linear SVM, logistic regression, and random forest baselines on the identical split, with hyperparameters chosen by cross-validation on the training data. The linear baselines reach 98.75\% at under 1\,ms per record on a CPU, outperforming the fine-tuned 9B model on this benchmark. We report this as a central finding: short, lexically distinctive benchmark narratives do not by themselves justify large decision models.
    \item \textbf{Structured Single-Pass Inference}: Because the model scores candidate classes directly without token generation, every output is a probability distribution over the fixed label set, so no free-text parsing is required. We did not benchmark the fine-tuned model's local latency (\Cref{sec:appendix:compute}).
    \item \textbf{Clause-Level Ablation Analysis}: We describe a Leave-One-Symptom-Out (LOSO) clause-ablation procedure and apply it to four illustrative narratives. In a psoriasis example, omitting the clause describing silvery scales lowers the predicted probability by 99.88 percentage points (from above 99.99\% to 0.12\%) and shifts the prediction to impetigo (99.59\%).
    \item \textbf{Stability to Secondary Clauses}: In the same examples, ablating other clauses changes the predicted probability by at most 0.67 percentage points, and by at most 0.02 points in the two multi-symptom narratives. This is consistent with the model relying on distinguishing features in these cases; a systematic analysis over the full test set remains future work.
    \item \textbf{Open-Source Clinical Decision Codebase}: To support reproducibility, all training pipelines, stratified data splits, NIH MedlinePlus diagnostic rubrics, counterfactual LOSO perturbation modules, and figure reproduction scripts are publicly released at \url{https://github.com/hrid0yyy/clef-clinical-triage}.
\end{itemize}

\section{Related Work}
\label{sec:related}

\subsection{Machine Learning on the Symptom2Disease Benchmark}
The \textit{Symptom2Disease} dataset \citep{barman2023symptom2disease} contains 1,200 short first-person symptom descriptions across 24 conditions (50 examples per class). Published studies have applied neural architectures to it. \citet{ijacsa2025disease} combined TF-IDF representations with an LSTM network, reporting 98.8\% accuracy on the 24-disease dataset. \citet{sbmufdsdd2026} developed FDSDD-LCEHOA, an ensemble incorporating BiLSTM, Temporal Convolutional Networks, Convolutional Autoencoders, and metaheuristic optimization, reporting 98.37\% accuracy on symptom-description data; the paper cites \textit{Symptom2Disease} but does not state explicitly which dataset was used for the reported result. These studies report high accuracy on this closed dataset but do not report clause-level attribution analyses.

\subsection{Generative Language Models in Clinical Decision Support}
Pretrained language models have shown strong capabilities on medical licensing examinations and clinical reasoning tasks \citep{singhal2023large}. On the Symptom2Disease benchmark, \citet{hasan2026clinllm} introduced CLIN-LLM, pairing a BioBERT classifier \citep{lee2020biobert} with Monte Carlo Dropout and an autoregressive FLAN-T5 model \citep{chung2024scaling} for treatment recommendations, reporting 98\% accuracy and F1 score, with ClinicalBERT \citep{alsentzer2019publicly} reaching an F1 score of 88.8\% as a baseline, and processing each case in under 2.5 seconds on an A100 GPU. CLIN-LLM's uncertainty estimation improves reliability, but maintaining two models and generating text step by step adds latency. Our approach uses a single decision foundation model that reaches lower accuracy (95.42\%) but requires no token generation.

\subsection{Explainability and Verification in Medical AI}
Applying machine learning in healthcare benefits from interpretability that is faithful to the model's actual behavior \citep{doshivelez2017rigorous}. In clinical natural language processing, feature attribution techniques generally fall into gradient-based methods and perturbation methods. Gradient-based estimators (such as Integrated Gradients \citep{sundararajan2017axiomatic}) measure sensitivity with respect to continuous input embeddings, but their attributions can depend strongly on the chosen baseline and do not align naturally with discrete semantic units in text. Perturbation methods \citep{ribeiro2016why} instead measure how the output changes when discrete units are removed. We use a Leave-One-Symptom-Out clause-ablation procedure to examine, in selected cases, whether predictions depend on clinically distinguishing symptoms rather than non-specific complaints.

\section{Methodology}
\label{sec:methods}

\subsection{Dataset Description and Clinical Scope}
\label{sec:methods:dataset}
We evaluate our approach on the \textit{Symptom2Disease} dataset \citep{barman2023symptom2disease}, a publicly available Kaggle dataset of 1,200 short first-person symptom descriptions distributed equally across 24 medical conditions (50 examples per class). The narratives describe subjective sensations, symptom onset, anatomical locations, and functional limitations (for instance, skin eruptions, joint stiffness, breathlessness, or episodic abdominal pain). The provenance of the narratives is not documented in detail, and we do not assume they were written by real patients. Across the 1,200 cases, narrative lengths range from 12 to 55 words, with a mean length of $30.73 \pm 6.71$ words and a mean character count of $171.88 \pm 35.48$ characters (range: 60 to 317 characters).

The 24 diagnostic categories can be grouped into four broad clinical domains. Infectious and acute febrile illnesses include dengue, malaria, typhoid, chicken pox, pneumonia, and common cold. Dermatological conditions comprise acne, impetigo, psoriasis, fungal infection, and drug reaction. Gastrointestinal and hepatic disorders consist of gastroesophageal reflux disease (GERD), peptic ulcer disease, jaundice, and dimorphic hemorrhoids\footnote{We retain the dataset's original label ``Dimorphic Hemorrhoids''. ``Dimorphic'' is not standard clinical terminology; the class corresponds to hemorrhoids (piles).}. The remaining nine conditions span respiratory (bronchial asthma), cardiovascular and vascular (hypertension, varicose veins), endocrine (diabetes), musculoskeletal (arthritis, cervical spondylosis), neurological (migraine), immunological (allergy), and urological (urinary tract infection) systems. Several conditions share constitutional symptoms such as fever, headache, body aches, and fatigue, so a model must use distinguishing features rather than non-specific indicators alone.

\subsection{Data Preprocessing and Decision Schema Formulation}
\label{sec:methods:preprocessing}
Unlike traditional machine learning pipelines that apply aggressive text normalization (such as stopword filtering, stemming, or punctuation removal), we intentionally retain the raw natural language text of each narrative. Punctuation marks (periods, commas, and semicolons) provide syntactic structure that delineates distinct symptom clauses, which is informative for attention mechanisms and necessary for clause-level perturbation analysis.

Data preparation involves three key steps:
\begin{enumerate}
    \item \textbf{Label Canonicalization}: Raw labels in the dataset contain minor casing discrepancies (for example, lowercase \texttt{diabetes} alongside title-cased \texttt{Psoriasis}). We standardize all 24 labels into a canonical, alphabetically sorted vocabulary $\mathcal{C} = \{c_1, c_2, \dots, c_{24}\}$.
    \item \textbf{Decision Schema Construction}: Each sample is structured into the native schema format required by the Clef-Flash processor:
    \begin{equation}
        \mathcal{R} = \left\{ \mathrm{state}: X, \; \mathrm{questions}: \big\{ \mathrm{disease}: (\mathrm{type}: \text{choice}, \; \mathrm{instructions}: I, \; \mathrm{options}: \mathcal{C}) \big\} \right\}
    \end{equation}
    where $X$ is the raw narrative and $I$ is the standardized instruction: \textit{``Which disease does this patient most likely have based on the reported symptoms?''}.
    \item \textbf{Ablation of External Diagnostic Definitions}: Although the schema head supports reference text criteria for each candidate class, we set criteria descriptions to \texttt{None} during model fine-tuning. This ensures the model learns true symptom-to-disease associations from training data, rather than relying on lexical matching against external textbook definitions.
\end{enumerate}

\subsection{Stratified Partitioning and Leakage Control}
\label{sec:methods:partition}
To evaluate the model on unseen cases, we apply a fixed, stratified 80/20 train-test partition across the 1,200 records. The training set contains 960 cases (40 per class) and the test set 240 cases (10 per class). All low-rank adapters and classification head parameters are trained only on the 960 training cases. No separate validation set was held out: the test split was evaluated after each epoch, and the checkpoint with the highest test accuracy (epoch 4) was retained. Reported test metrics are therefore subject to some optimistic bias from checkpoint selection. All results come from a single training run (random seed 42).

\subsection{Classical Baselines on the Same Split}
\label{sec:methods:baselines}
Because published results on \textit{Symptom2Disease} use different splits, we trained three classical baselines on the same 960 training cases and evaluated them on the same 240 test cases. Each uses unigram and bigram TF-IDF features (sublinear term frequency) followed by a linear support vector machine \citep{cortes1995support}, a multinomial logistic regression, or a random forest \citep{breiman2001random}. Hyperparameters (regularization strength $C$ for the linear models; number of trees and features per split for the random forest) were selected by stratified 5-fold cross-validation on the training split only, and each final model was evaluated once on the test split. Latency is the mean single-record inference time (vectorization and prediction) over the 240 test records on a laptop CPU (AMD Ryzen 7 250). The script and outputs are included in the code release (\texttt{tools/run\_classical\_baselines.py}).

\subsection{Decision Foundation Model Architecture: Clef-Flash}
The Clef-Flash model \citep{cloudflare2026clef} connects a language backbone (Qwen3.5-9B, 32 layers, hidden size $d = 4096$) \citep{qwen35model} to a small transformer scoring head, the \textbf{JointSchemaHead}. The description below follows the released implementation (\texttt{joint\_schema\_model.py}) and head configuration.

The serialized schema (narrative $X$, instruction, and the $K = 24$ option strings $\mathcal{C} = \{c_1, \dots, c_K\}$) is encoded by the backbone into per-token hidden states $H \in \mathbb{R}^{T \times d}$. The head projects $H$ into a token memory of width 1024. Each option $c_k$ is represented by a query vector $o_k$ formed from three projected terms: the mean hidden state over the option's tokens, the mean of the backbone's output-embedding rows for those tokens (a lexical vector $\ell_k$), and the mean hidden state over the question span $q$. Two evidence-routing layers let the option queries cross-attend to the token memory, and a 4-layer transformer decoder (width 1024, 16 heads, feed-forward size 4096) produces a field vector $f$ for the question, combining the question, an attention-weighted summary of the routed options, and the final-token state $g$. The logit for option $c_k$ is
\begin{equation}
    z_k = \underbrace{s_p \, \cos\!\big(\ell_k,\; q + g\big)}_{\text{lexical prior}} \;+\; \sigma(\gamma)\,\Big[\, s_j \cos\!\big(f, \tilde{o}_k\big) + \mathrm{MLP}\big([f, \tilde{o}_k, f \odot \tilde{o}_k, |f - \tilde{o}_k|]\big) \Big],
\end{equation}
where $\tilde{o}_k$ is the normalized routed option vector, $s_p$ and $s_j$ are learned logit scales (each capped at 100), and $\sigma(\gamma)$ is a learned gate initialized at 0.5. Class probabilities follow from a softmax over the $K$ options:
\begin{equation}
    P(Y = c_k \mid X) = \frac{\exp(z_k)}{\sum_{j=1}^{K} \exp(z_j)}
\end{equation}
Because all $K$ candidates are scored in one forward pass, the model needs no sequential decoding loop and cannot produce malformed output.

\subsection{Parameter-Efficient Fine-Tuning Formulation}
To tune the 9-billion parameter model within local computational constraints, we apply 4-bit Quantized Low-Rank Adaptation (QLoRA) \citep{dettmers2023qlora, hu2022lora}. Backbone weights $W_0 \in \mathbb{R}^{d \times d}$ are held frozen and stored in 4-bit NormalFloat (NF4) representation with double quantization. Trainable low-rank decomposition matrices $A \in \mathbb{R}^{r \times d}$ and $B \in \mathbb{R}^{d \times r}$ with rank $r = 16$ and scaling factor $\alpha = 32$ are injected into the attention projection modules:
\begin{equation}
    W = W_0 + \frac{\alpha}{r} B A
\end{equation}
where adapters target the query ($q\_proj$), key ($k\_proj$), value ($v\_proj$), and output ($o\_proj$) linear projections.

In parallel, all parameters within the JointSchemaHead, including normalization layers, projection weights, attention and decoder layers, and logit scales, remain fully trainable in \texttt{bfloat16} precision. To balance gradient dynamics across the frozen-quantized backbone and the classification head, we employ differential learning rates: a base rate of $\eta_{\mathrm{backbone}} = 1.0 \times 10^{-4}$ for the low-rank attention adapters and an elevated rate of $\eta_{\mathrm{head}} = 3.0 \times 10^{-4}$ for the JointSchemaHead to foster rapid and stable convergence.

Training optimizes the cross-entropy objective over true label $y^*$:
\begin{equation}
    \mathcal{L}_{\mathrm{CE}} = - \log P(Y = y^* \mid X)
\end{equation}
Optimization uses AdamW \citep{loshchilov2019decoupled} ($\beta_1=0.9, \beta_2=0.999$, weight decay $0.01$) in PyTorch \citep{paszke2019pytorch} with Hugging Face PEFT \citep{peft}. We use a linear warmup over the first $10\%$ of steps followed by cosine decay across 5 epochs. With a per-device batch size of 4 and 4 gradient accumulation steps, the effective batch size is 16, and gradients are clipped to a norm of 1.0. Fine-tuning, including per-epoch test evaluation, took 34.6 minutes on one NVIDIA GeForce RTX 4090 GPU (24GB VRAM). The trainable adapter checkpoint contains 186 parameter tensors occupying 247.31\,MB on disk.

\subsection{Counterfactual Perturbation and Clinical Auditability}
\label{sec:methods:counterfactual}
To examine the decision basis of the fine-tuned model, and whether its predictions depend on clinically distinguishing evidence rather than spurious lexical correlations, we use a Leave-One-Symptom-Out (LOSO) clause-ablation procedure.

\paragraph{Clause Decomposition and Perturbation Formulation.}
Each patient narrative $X$ is partitioned into an ordered set of disjoint clinical clauses $S = \{s_1, s_2, \dots, s_M\}$, where each clause corresponds to an individual reported symptom, complication, or anatomical descriptor. For each clause $s_m \in S$, we construct an ablated counterfactual narrative $X_{\setminus s_m} = X \setminus \{s_m\}$ by removing clause $s_m$ while preserving the grammatical coherence and punctuation of the surrounding context.

\paragraph{Marginal Confidence and Relative Sensitivity.}
Let $P(Y = y^* \mid X)$ denote the model's predicted probability for the target disease $y^*$ under the complete narrative. The marginal confidence drop $\Delta P(s_m)$ and relative drop percentage $\mathrm{Drop}(s_m)$ induced by omitting clause $s_m$ are defined as:
\begin{align}
    \Delta P(s_m) &= P(Y = y^* \mid X) - P(Y = y^* \mid X_{\setminus s_m}) \\
    \mathrm{Drop}(s_m) &= \frac{\Delta P(s_m)}{P(Y = y^* \mid X)} \times 100\%
\end{align}

\paragraph{Counterfactual Decision Flips and Differential Reallocation.}
A counterfactual decision flip occurs when omitting clause $s_m$ causes the primary predicted candidate to shift away from the ground-truth condition. The post-ablation primary diagnosis is given by:
\begin{equation}
    \hat{y}_{\setminus s_m} = \operatorname{argmax}_{k \in \{1, \dots, K\}} P(Y = c_k \mid X_{\setminus s_m})
\end{equation}
with a decision flip defined by the condition $\hat{y}_{\setminus s_m} \neq y^*$. When a flip occurs, we examine the redistribution of probability mass across alternative candidate diagnoses. The differential reallocation mass toward the new primary diagnosis $\hat{y}_{\setminus s_m}$ is defined as:
\begin{equation}
    \mathcal{M}_{\mathrm{alt}}(s_m) = P(Y = \hat{y}_{\setminus s_m} \mid X_{\setminus s_m})
\end{equation}
A clinically faithful model should not reallocate probability mass to an arbitrary or unrelated disorder. Instead, the counterfactual shift should align with recognized medical differentials that share the remaining symptom profile.

\paragraph{Diagnostic Invariance Bound for Secondary Symptoms.}
Conversely, when removing non-defining or secondary symptoms $s_{\mathrm{sec}}$, a reliable diagnostic model should exhibit minimal sensitivity, satisfying an invariance bound:
\begin{equation}
    \Delta P(s_{\mathrm{sec}}) \le \epsilon_{\mathrm{sec}}
\end{equation}
where $\epsilon_{\mathrm{sec}} \ll 1\%$, which would indicate that the prediction is stable to incidental or non-discriminative complaints.

\section{Experimental Results}
\label{sec:results}

\subsection{Comparison with Baselines}
\Cref{tab:scorecard} compares the fine-tuned Clef-Flash model with classical baselines trained and evaluated on the same split (Panel A) and with results reported in the literature on other splits of \textit{Symptom2Disease} (Panel B).

\begin{table}[htbp]
\centering
\small
\caption{\textbf{Comparative Performance on Symptom2Disease.} Panel A: all models use the same 960/240 split; classical baselines were tuned by 5-fold cross-validation on the training split and evaluated once on the test split (95\% Wilson CIs in brackets). Classical latencies are single-record CPU inference (AMD Ryzen 7 250); zero-shot API latencies are mean per-request times to the remote Cloudflare Workers AI endpoint, including network round trips. Zero-shot rows use bare option names without disease criteria, matching the fine-tuning setup. The fine-tuned model's local latency was not measured. Panel B: results as reported by the original authors on their own splits and hardware; they are not directly comparable to Panel A. ``--'' means not reported. $^\dagger$The paper cites \textit{Symptom2Disease} but does not state explicitly that the reported result uses it. $^\ddagger$Averaging method not stated.}
\label{tab:scorecard}
\resizebox{\textwidth}{!}{%
\begin{tabular}{llcccc}
\toprule
\textbf{Model} & \textbf{Architecture} & \textbf{Accuracy [95\% CI]} & \textbf{Macro F1} & \textbf{Output Tok.} & \textbf{Latency} \\
\midrule
\multicolumn{6}{l}{\textbf{Panel A: Same split (this work)}} \\
\midrule
TF-IDF + Linear SVM & Uni/bigram TF-IDF, linear SVM & \textbf{98.75\%} [96.39--99.57] & \textbf{0.9876} & 0 & 0.7\,ms (CPU) \\
TF-IDF + Logistic Regression & Uni/bigram TF-IDF, logistic regression & \textbf{98.75\%} [96.39--99.57] & \textbf{0.9876} & 0 & 0.8\,ms (CPU) \\
TF-IDF + Random Forest & Uni/bigram TF-IDF, 500 trees & 96.67\% [93.56--98.30] & 0.9665 & 0 & 45\,ms (CPU) \\
Clef (27B), zero-shot API \citep{clef27bmodel} & Qwen3.8-27B + JointSchemaHead & 66.25\% [60.05--71.93] & 0.6186 & 0 & 0.92\,s (API) \\
Clef-Flash (9B), zero-shot API \citep{cloudflare2026clef} & Qwen3.5-9B + JointSchemaHead & 62.92\% [56.65--68.78] & 0.5985 & 0 & 0.80\,s (API) \\
Clef-Flash (9B), zero-shot local \citep{cloudflare2026clef} & Qwen3.5-9B (4-bit) + JointSchemaHead & 65.00\% [58.77--70.76] & -- & 0 & -- \\
Fine-tuned Clef-Flash (ours) & Qwen3.5-9B QLoRA + JointSchemaHead & 95.42\% [91.98--97.42] & 0.9551 & 0 & not measured \\
\midrule
\multicolumn{6}{l}{\textbf{Panel B: Reported in the literature (different splits)}} \\
\midrule
Al-qarni and Algarni \citep{ijacsa2025disease} & TF-IDF + LSTM & 98.8\% & -- & 0 & -- \\
FDSDD-LCEHOA \citep{sbmufdsdd2026} & Ensemble (BiLSTM+TCN+CAE) & 98.37\%$^\dagger$ & -- & 0 & -- \\
CLIN-LLM \citep{hasan2026clinllm} & BioBERT + MC Dropout + FLAN-T5 & 98\% & 0.98$^\ddagger$ & generated & $<2.5$\,s (A100) \\
ClinicalBERT (in \citep{hasan2026clinllm}) \citep{alsentzer2019publicly} & ClinicalBERT classifier & -- & 0.888$^\ddagger$ & 0 & -- \\
\bottomrule
\end{tabular}%
}
\end{table}

The fine-tuned Clef-Flash model achieves \textbf{95.42\% accuracy} (229 of 240 test cases; Wilson 95\% CI 91.98--97.42\%) with a \textbf{macro F1 score of 0.9551}. This is a 30.42 percentage point gain over the same unadapted weights evaluated locally (65.00\%) and a 32.50 point gain over the zero-shot Cloudflare Workers AI endpoint (62.92\%). The two zero-shot figures come from different serving setups (local 4-bit weights versus the remote API), and we report both; per-class zero-shot results in \Cref{tab:classwise} use the API run, the only zero-shot run with per-class outputs saved. When the MedlinePlus criteria summaries (\Cref{tab:app_dataset}) were added to the prompt, zero-shot API accuracy was 61.67\% for Clef-Flash and 67.92\% for the 27B Clef model, so the criteria did not consistently help.

On the same split, however, simple TF-IDF baselines are more accurate. A linear SVM and a logistic regression classifier over unigram and bigram TF-IDF features both reach 98.75\% (237 of 240; macro F1 0.9876), and a TF-IDF random forest reaches 96.67\% (95.00 to 97.08\% across five random seeds). The fine-tuned model makes 11 errors against 3 for the linear models. The confidence intervals overlap only marginally, but because per-case predictions of the fine-tuned model were not retained, we could not run a paired significance test. The linear baselines also run in under 1\,ms per record on a laptop CPU, whereas the decision model requires a forward pass through a 9B-parameter backbone on a GPU. On this benchmark, therefore, the fine-tuned decision model offers no accuracy or cost advantage over a well-tuned linear classifier. Its potential value lies elsewhere: it starts from a general-purpose language backbone, scores a schema rather than a fixed feature vocabulary, and may transfer better to longer and more varied clinical text than short, lexically distinctive benchmark narratives. This study does not test that.

Relative to generative pipelines such as CLIN-LLM, which generates treatment text and reports under 2.5\,s per case on an A100 GPU, the decision model needs only one forward pass and no decoding; every output is a probability distribution over the fixed label set, so no free-text parsing is required. We did not measure the fine-tuned model's local latency, so we make no quantitative speed claim. The literature accuracies in Panel B (98 to 98.8\%) are consistent with our same-split linear baselines but come from different splits and are not directly comparable.

\subsection{Training Dynamics and Convergence}
\begin{figure*}[t]
\centering
\includegraphics[width=\textwidth]{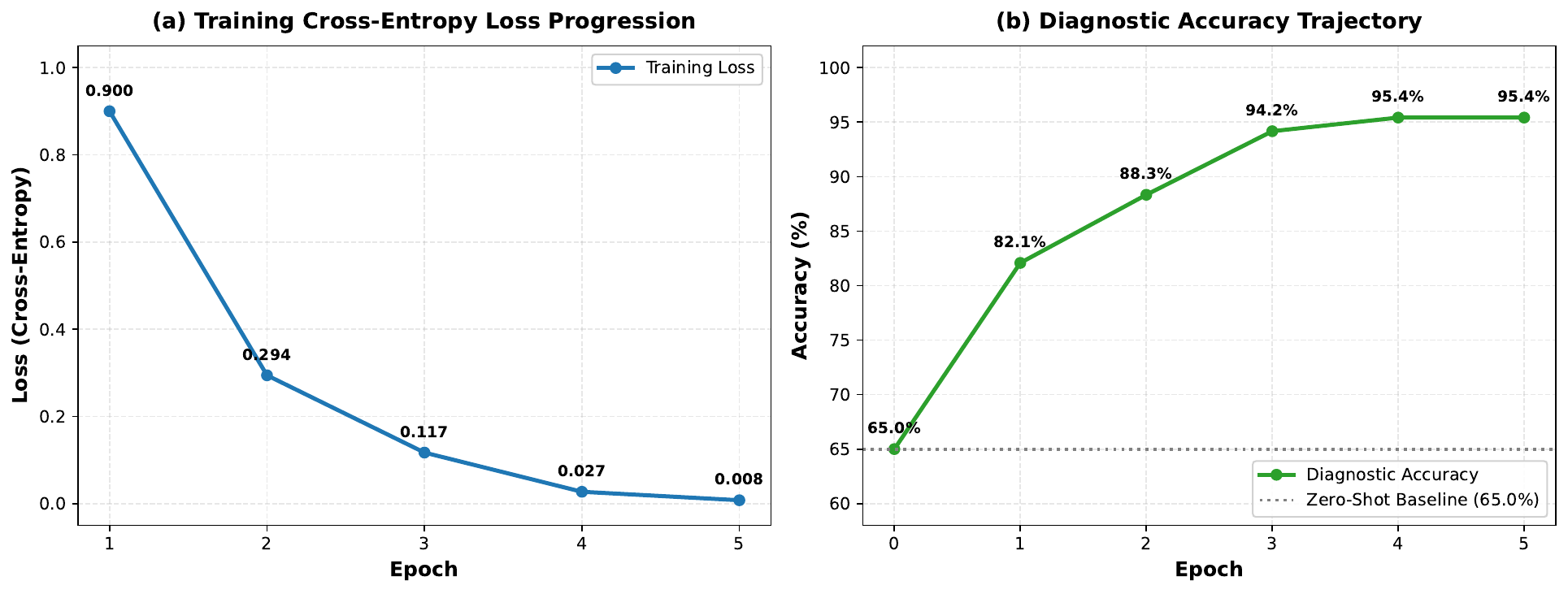}
\caption{\textbf{Training loss and test-set accuracy of Clef-Flash across 5 epochs.} 
(A) Cross-entropy training loss progression across epochs, showing rapid and monotonic convergence from 0.900 down to 0.008. 
(B) Test-set accuracy from the zero-shot baseline (65.00\%) to 95.42\% across 5 training epochs. The test split was also used for checkpoint selection (\Cref{sec:methods:partition}).}
\label{fig:training_dynamics}
\end{figure*}

\Cref{fig:training_dynamics} presents the training loss and accuracy progression across the 5 training epochs.

Training cross-entropy loss decreased from 0.900 in epoch 1 to 0.294 in epoch 2, 0.117 in epoch 3, 0.027 in epoch 4, and 0.008 by epoch 5. Test accuracy rose from the zero-shot baseline of 65.00\% to 82.08\% in epoch 1, 88.33\% in epoch 2, 94.17\% in epoch 3, and 95.42\% in epoch 4, remaining at 95.42\% in epoch 5. Because the test split was monitored during training, these trajectories describe the run rather than unbiased generalization estimates.

\subsection{Error Analysis and Confusion Matrix}
\begin{figure*}[t]
\centering
\includegraphics[width=0.88\textwidth]{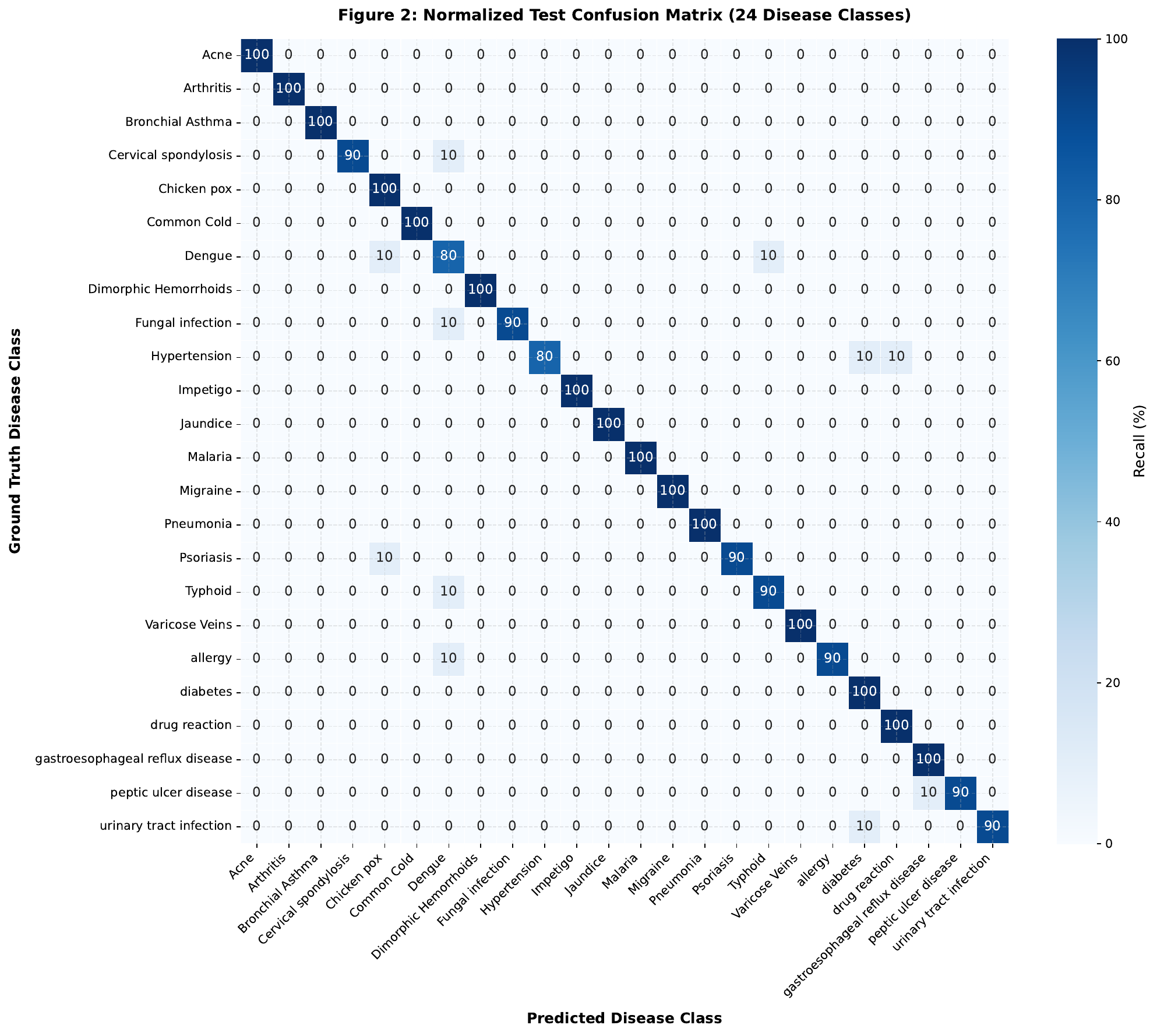}
\caption{\textbf{Normalized $24 \times 24$ confusion matrix on the 240 held-out test cases.} 15 of the 24 conditions achieved 100\% recall (10 out of 10 cases correct), with 11 classes reaching 100\% precision and recall. Six of the 11 misclassifications involve dengue, either as the true or the predicted class.}
\label{fig:confusion_matrix}
\end{figure*}

To examine prediction distributions across specific classes, we computed the normalized $24 \times 24$ confusion matrix on the 240 test cases (\Cref{fig:confusion_matrix}).

Across the 24 disease classes, \textbf{15 conditions achieved 100\% recall} (all 10 test cases correctly identified): \textit{Acne, Arthritis, Bronchial Asthma, Chicken pox, Common Cold, Diabetes, Dimorphic Hemorrhoids, Drug reaction, GERD, Impetigo, Jaundice, Malaria, Migraine, Pneumonia}, and \textit{Varicose Veins}. Among these, 11 conditions achieved perfect classification with both 100\% precision and 100\% recall (Acne, Arthritis, Bronchial Asthma, Common Cold, Dimorphic Hemorrhoids, Impetigo, Jaundice, Malaria, Migraine, Pneumonia, and Varicose Veins).

Across the 240-case test set, 11 misclassifications occurred (95.42\% accuracy); \Cref{tab:app_errors} lists each by true and predicted class. Six involve dengue: two dengue cases were predicted as chicken pox and typhoid, and four cases from cervical spondylosis, fungal infection, typhoid, and allergy were predicted as dengue. This is consistent with the model over-weighting non-specific fever, rash, and body-ache descriptions toward dengue, which has the lowest precision of any class (0.667).

The remaining errors are urinary tract infection predicted as diabetes, psoriasis as chicken pox, peptic ulcer disease as GERD, and two hypertension cases predicted as diabetes and drug reaction. Several of these pairs share presenting symptoms (for example, urinary frequency in urinary tract infection and diabetes, or epigastric burning in peptic ulcer disease and GERD). However, the errors were not reviewed by clinicians, and some (such as hypertension predicted as drug reaction) have no obvious clinical explanation.

\subsection{Classwise Performance Gains}
\begin{figure*}[t]
\centering
\includegraphics[width=\textwidth]{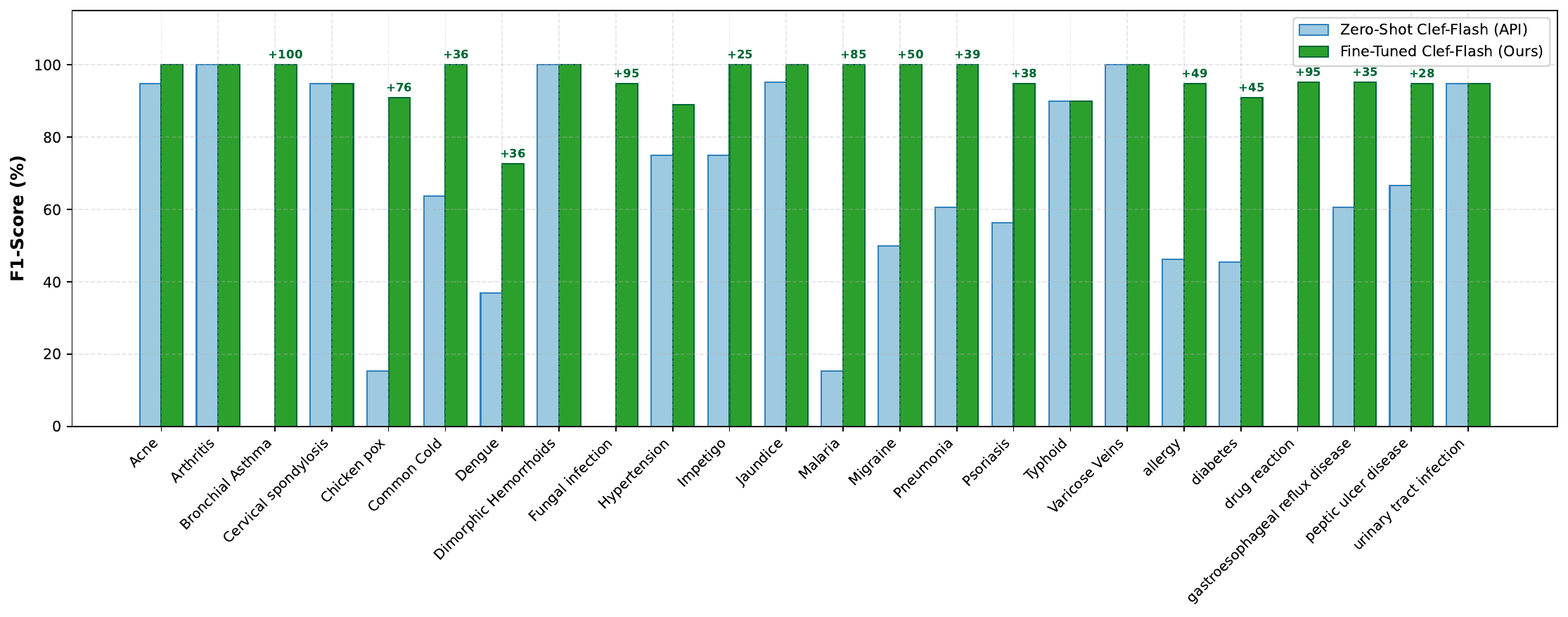}
\caption{\textbf{Class-by-class F1-score comparison between zero-shot Clef-Flash (remote API, bare option names) and the fine-tuned model.} The largest gains occur for conditions with low zero-shot recognition, including Bronchial Asthma ($+1.00$), Drug Reaction ($+0.95$), Fungal Infection ($+0.95$), Malaria ($+0.85$), and Chicken Pox ($+0.76$).}
\label{fig:classwise}
\end{figure*}

\Cref{tab:classwise} and \Cref{fig:classwise} summarize precision, recall, and F1 scores across all 24 conditions alongside improvements over the zero-shot baseline.

\begin{table}[htbp]
\centering
\small
\caption{\textbf{Classwise Diagnostic Performance on Held-Out Test Set (240 Cases).} Zero-Shot is the F1 of Clef-Flash through the remote API with bare option names; Gain is the fine-tuned F1 minus the zero-shot F1.}
\label{tab:classwise}
\resizebox{\textwidth}{!}{%
\begin{tabular}{lccccc|lccccc}
\toprule
\textbf{Disease Class} & \textbf{Zero-Shot} & \textbf{Prec.} & \textbf{Rec.} & \textbf{F1} & \textbf{Gain ($\Delta$)} & \textbf{Disease Class} & \textbf{Zero-Shot} & \textbf{Prec.} & \textbf{Rec.} & \textbf{F1} & \textbf{Gain ($\Delta$)} \\
\midrule
Acne & 0.9474 & 1.0000 & 1.0000 & 1.0000 & +0.0526 & Malaria & 0.1538 & 1.0000 & 1.0000 & 1.0000 & +0.8462 \\
Arthritis & 1.0000 & 1.0000 & 1.0000 & 1.0000 & +0.0000 & Migraine & 0.5000 & 1.0000 & 1.0000 & 1.0000 & +0.5000 \\
Bronchial Asthma & 0.0000 & 1.0000 & 1.0000 & 1.0000 & +1.0000 & Pneumonia & 0.6061 & 1.0000 & 1.0000 & 1.0000 & +0.3939 \\
Cervical spondylosis & 0.9474 & 1.0000 & 0.9000 & 0.9474 & +0.0000 & Psoriasis & 0.5625 & 1.0000 & 0.9000 & 0.9474 & +0.3849 \\
Chicken pox & 0.1538 & 0.8333 & 1.0000 & 0.9091 & +0.7552 & Typhoid & 0.9000 & 0.9000 & 0.9000 & 0.9000 & +0.0000 \\
Common Cold & 0.6364 & 1.0000 & 1.0000 & 1.0000 & +0.3636 & Varicose Veins & 1.0000 & 1.0000 & 1.0000 & 1.0000 & +0.0000 \\
Dengue & 0.3684 & 0.6667 & 0.8000 & 0.7273 & +0.3589 & Allergy & 0.4615 & 1.0000 & 0.9000 & 0.9474 & +0.4858 \\
Dimorphic Hemorrhoids & 1.0000 & 1.0000 & 1.0000 & 1.0000 & +0.0000 & Diabetes & 0.4545 & 0.8333 & 1.0000 & 0.9091 & +0.4545 \\
Fungal infection & 0.0000 & 1.0000 & 0.9000 & 0.9474 & +0.9474 & Drug reaction & 0.0000 & 0.9091 & 1.0000 & 0.9524 & +0.9524 \\
Hypertension & 0.7500 & 1.0000 & 0.8000 & 0.8889 & +0.1389 & GERD & 0.6061 & 0.9091 & 1.0000 & 0.9524 & +0.3463 \\
Impetigo & 0.7500 & 1.0000 & 1.0000 & 1.0000 & +0.2500 & Peptic ulcer disease & 0.6667 & 1.0000 & 0.9000 & 0.9474 & +0.2807 \\
Jaundice & 0.9524 & 1.0000 & 1.0000 & 1.0000 & +0.0476 & UTI & 0.9474 & 1.0000 & 0.9000 & 0.9474 & +0.0000 \\
\midrule
\multicolumn{6}{l|}{\textbf{Macro Average (All 24 Classes)}} & \multicolumn{6}{l}{\textbf{Precision: 0.9605 \quad Recall: 0.9542 \quad F1: 0.9551 \quad Gain: +0.3566}} \\
\bottomrule
\end{tabular}%
}
\end{table}

Zero-shot per-class scores come from the Clef-Flash API run with bare option names (62.92\% accuracy). Several categories with low zero-shot scores improved markedly with adaptation, including bronchial asthma ($0.0000 \to 1.0000$), drug reaction ($0.0000 \to 0.9524$), fungal infection ($0.0000 \to 0.9474$), malaria ($0.1538 \to 1.0000$), chicken pox ($0.1538 \to 0.9091$), and allergy ($0.4615 \to 0.9474$). No class's F1 decreased; six (arthritis, cervical spondylosis, dimorphic hemorrhoids, typhoid, urinary tract infection, and varicose veins) were unchanged. Across all 24 classes, adaptation lifted the macro F1 score from 0.5985 to 0.9551, a net gain of +0.3566.

\section{Clause-Level Ablation Analysis}
\label{sec:xai}

\subsection{Quantitative Hallmark versus Secondary Clause Ablation}
\label{sec:xai:quantitative}
\begin{figure*}[t]
\centering
\includegraphics[width=0.96\textwidth,keepaspectratio]{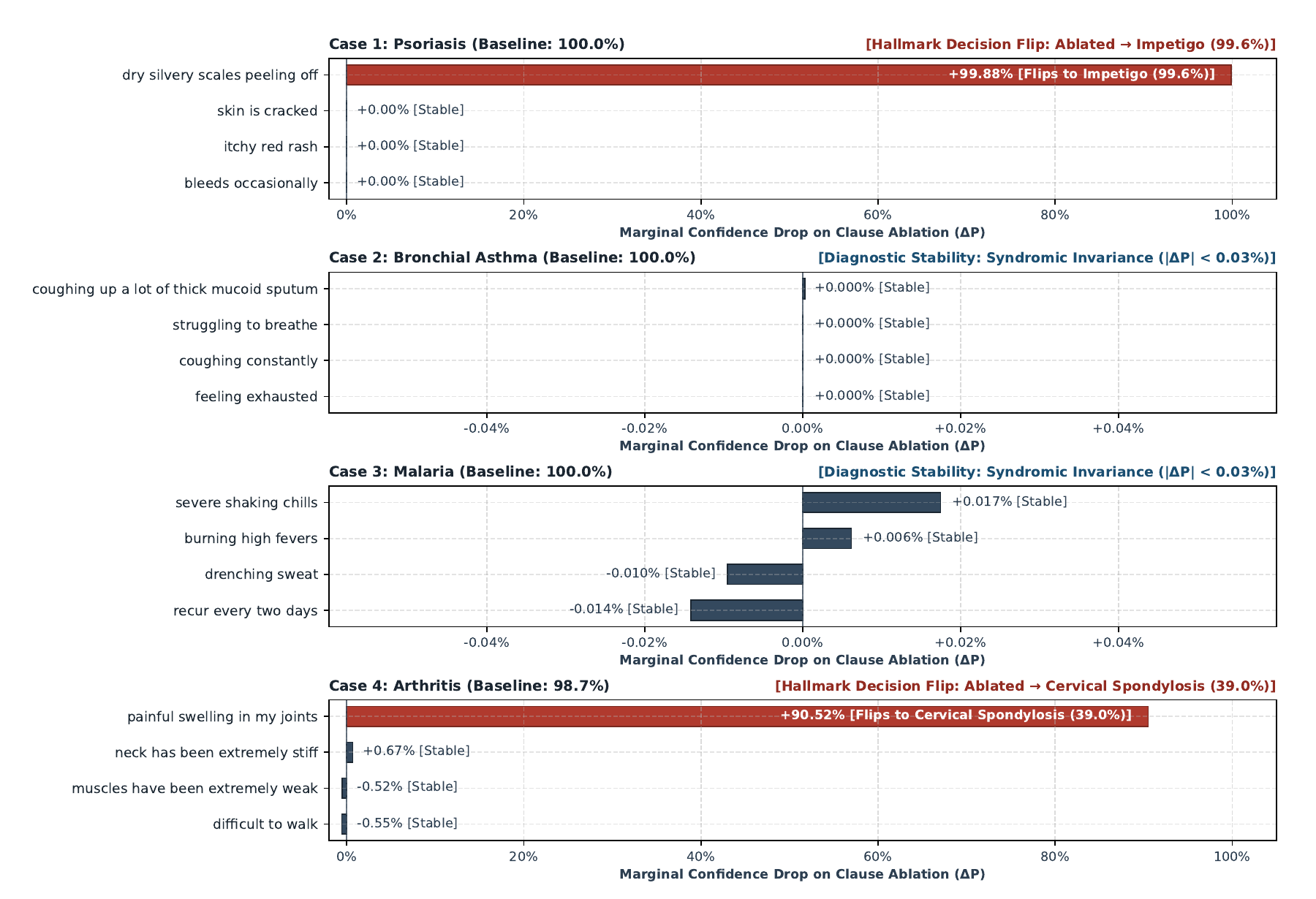}
\caption{\textbf{Leave-One-Symptom-Out (LOSO) clause ablation on four illustrative, author-written narratives.} Horizontal bars show the drop in predicted probability ($\Delta P$) when individual symptom clauses are removed from the input narrative. Red bars mark clauses whose removal flips the prediction (Psoriasis to Impetigo upon removal of ``dry silvery scales peeling off'', and Arthritis to Cervical Spondylosis upon removal of ``painful swelling in my joints''). In the two multi-symptom narratives (Bronchial Asthma and Malaria), removing any single clause changes the probability by at most 0.02 percentage points.}
\label{fig:perturbation}
\end{figure*}

To examine what the fine-tuned model relies on, we apply the Leave-One-Symptom-Out (LOSO) procedure to four narratives chosen for illustration: psoriasis, arthritis, bronchial asthma, and malaria. These narratives were written by the author to contain clearly separable clauses and are not drawn from the dataset. The clause categories (hallmark, secondary, non-defining, syndromic) were assigned by the author and have not been validated by clinicians. The results are therefore case studies, not a systematic evaluation. \Cref{tab:xai} and \Cref{fig:perturbation} report the confidence drops and post-ablation predictions.

\begin{table}[htbp]
\centering
\small
\caption{\textbf{Symptom Clause Ablation and Post-Ablation Predictions (four illustrative, author-written narratives; clause categories assigned by the author).} Panel A: in the psoriasis and arthritis narratives, removing the clause labeled as the hallmark lowers the target probability by at least 90 percentage points and changes the predicted class. Panel B: in the bronchial asthma and malaria narratives, which contain several corroborating clauses, removing any single clause changes the probability by at most 0.02 points.}
\label{tab:xai}
\resizebox{\textwidth}{!}{%
\begin{tabular}{llccccc}
\toprule
\textbf{Target Disease} & \textbf{Ablated Symptom Expression} & \textbf{Clinical Category} & \textbf{Base $P$} & \textbf{Post $P$} & \textbf{Drop ($\Delta$)} & \textbf{Post-Ablation Diagnosis} \\
\midrule
\multicolumn{7}{l}{\textbf{Panel A: Single-Anchor Presentations (Decision Flips)}} \\
\midrule
Psoriasis & dry silvery scales peeling off & HALLMARK DEFINING & 100.00\% & 0.12\% & \textbf{99.88\%} & Impetigo (99.59\%) \\
Psoriasis & skin is cracked & SECONDARY COMPLICATION & 100.00\% & 100.00\% & 0.00\% & Psoriasis (Stable) \\
Psoriasis & itchy red rash & NON-DEFINING / GENERAL & 100.00\% & 100.00\% & 0.00\% & Psoriasis (Stable) \\
Psoriasis & bleeds occasionally & NON-DEFINING / GENERAL & 100.00\% & 100.00\% & -0.00\% & Psoriasis (Stable) \\
\midrule
Arthritis & painful swelling in my joints & HALLMARK DEFINING & 98.69\% & 8.18\% & \textbf{90.52\%} & Cervical spondylosis (39.01\%) \\
Arthritis & neck has been extremely stiff & SECONDARY COMPLICATION & 98.69\% & 98.02\% & 0.67\% & Arthritis (Stable) \\
Arthritis & muscles have been extremely weak & NON-DEFINING / GENERAL & 98.69\% & 99.21\% & -0.52\% & Arthritis (Stable) \\
Arthritis & difficult to walk & NON-DEFINING / GENERAL & 98.69\% & 99.25\% & -0.55\% & Arthritis (Stable) \\
\midrule
\multicolumn{7}{l}{\textbf{Panel B: Multi-Symptom Presentations (Syndromic Redundancy)}} \\
\midrule
Bronchial Asthma & coughing up a lot of thick mucoid sputum & CARDINAL SYNDROMIC & 100.00\% & 100.00\% & 0.00\% & Bronchial Asthma (Stable) \\
Bronchial Asthma & struggling to breathe & CARDINAL SYNDROMIC & 100.00\% & 100.00\% & 0.00\% & Bronchial Asthma (Stable) \\
Bronchial Asthma & coughing constantly & CARDINAL SYNDROMIC & 100.00\% & 100.00\% & 0.00\% & Bronchial Asthma (Stable) \\
Bronchial Asthma & feeling exhausted & NON-DEFINING / GENERAL & 100.00\% & 100.00\% & -0.00\% & Bronchial Asthma (Stable) \\
\midrule
Malaria & severe shaking chills & CARDINAL SYNDROMIC & 99.98\% & 99.96\% & 0.02\% & Malaria (Stable) \\
Malaria & burning high fevers & CARDINAL SYNDROMIC & 99.98\% & 99.97\% & 0.01\% & Malaria (Stable) \\
Malaria & drenching sweat & CARDINAL SYNDROMIC & 99.98\% & 99.98\% & -0.01\% & Malaria (Stable) \\
Malaria & recur every two days & CARDINAL SYNDROMIC & 99.98\% & 99.99\% & -0.01\% & Malaria (Stable) \\
\bottomrule
\end{tabular}%
}
\end{table}

In these examples, ablation separates narratives that rest on a single distinguishing clause from narratives with several corroborating clauses. Removing the hallmark clause lowers the predicted probability by 99.88 percentage points in the psoriasis narrative (from above 99.99\% to 0.12\%) and by 90.52 points in the arthritis narrative (from 98.69\% to 8.18\%, a 91.72\% relative drop), and in both cases the predicted class changes. In the bronchial asthma and malaria narratives, removing any single clause changes the probability by at most 0.02 points, consistent with the remaining clauses still supporting the diagnosis.

\subsection{Clinical Plausibility of Differential Reallocations}
\label{sec:xai:differential}
In both flipped cases, the new prediction can be compared with the differential a clinician might consider for the remaining text.

In the psoriasis case, the complete narrative describes an itchy red rash with cracked, bleeding skin and peeling silvery scales. When the clause \textit{``dry silvery scales peeling off''} is removed, the remaining description is an itchy red rash that is cracked and bleeding. Without the characteristic silvery scale, impetigo is a reasonable differential for such a description, and the model assigns it 99.59\% of the probability mass. The very high confidence on a fairly non-specific residual description is itself notable, since a clinician would likely keep several possibilities open.

In the arthritis case, the narrative describes painful joint swelling with neck stiffness, muscle weakness, and difficulty walking. When \textit{``painful swelling in my joints''} is removed, the remaining narrative describes neck stiffness, generalized weakness, and difficulty walking. The model's top prediction becomes Cervical Spondylosis (39.01\%), with drug reaction close behind (36.64\%). Cervical spondylosis is a reasonable reading of neck stiffness with weakness; drug reaction is not an obvious differential. The split probability shows that the model is uncertain on this residual narrative rather than moving to a single coherent alternative.

\subsection{Stability to Secondary Clauses}
\label{sec:xai:invariance}
Removing secondary complications, non-specific complaints, or individual components of multi-symptom presentations changes the output distribution little in these examples.

In the psoriasis narrative, removing \textit{``skin is cracked''} lowers the probability by less than 0.001 percentage points, and removing \textit{``itchy red rash''} produces no measurable change. In the arthritis narrative, removing \textit{``neck has been extremely stiff''} lowers the probability by 0.67 points (from 98.69\% to 98.02\%), and arthritis remains the prediction.

In the bronchial asthma narrative, which contains breathlessness (\textit{``struggling to breathe''}), persistent cough (\textit{``coughing constantly''}), and sputum (\textit{``coughing up a lot of thick mucoid sputum''}), removing any one of these clauses changes the probability by less than 0.001 points. In the malaria narrative, which includes periodic episodes (\textit{``recur every two days''}), shaking chills (\textit{``severe shaking chills''}), and high fevers (\textit{``burning high fevers''}), removing any single clause changes the probability by at most 0.02 points ($|\Delta P| \le 0.018$).

Taken together, in these four examples the model was sensitive to removing a single distinguishing clause and stable when redundant evidence remained. Because the narratives were written and selected by the author and the clause categories were not validated by clinicians, this is suggestive rather than conclusive. Supporting general claims about the model's decision basis would require LOSO analysis over the full test set with clinician-annotated clause categories.

\section{Discussion and Limitations}
\label{sec:discussion}

\subsection{Clinical Implications of Zero-Generation Architectures}
By scoring candidate diagnoses directly through the JointSchemaHead, the system always returns a probability distribution over the fixed label set rather than free text. This removes free-text hallucination and output-parsing failures as failure modes, although the model can still be confidently wrong, and its probabilities have not been evaluated for calibration. Its outputs map directly onto structured fields, which may simplify integration with electronic health record systems, although we have not tested such integration.

A single forward pass with no decoding is, in principle, well suited to interactive use in high-volume intake settings, although we did not measure local latency. On this benchmark, TF-IDF linear classifiers trained on the same split are both more accurate (98.75\% versus 95.42\%) and far cheaper, running in under 1\,ms per record on a CPU. The practical case for a 9B-parameter model therefore rests on its potential to handle more varied and realistic language than Symptom2Disease contains, which this study does not test. Clause-level ablation provides one inspectable explanation for individual predictions, which may support clinician review and algorithmic accountability; we did not evaluate whether it satisfies any specific regulatory requirement.

\subsection{Limitations and Future Directions}
Several methodological and practical limitations should be considered:
\begin{enumerate}
    \item \textbf{Corpus Scale, Provenance, and Labels}: The \textit{Symptom2Disease} dataset comprises 1,200 records (50 per condition), was chosen partly because of single-GPU (RTX 4090, 24GB) compute constraints, and is a public Kaggle dataset whose narratives have undocumented provenance and may not reflect how real patients describe their symptoms. Some labels are non-standard (``Dimorphic Hemorrhoids''). Evaluating on larger, real clinical corpora is needed.
    \item \textbf{Exploration of Alternative Fine-Tuning Regimes}: Our investigation focused on 4-bit Quantized Low-Rank Adaptation (QLoRA) targeting attention projection layers ($q, k, v, o$). Future work could systematically explore alternative parameter-efficient adaptation paradigms, such as Weight-Decomposed Low-Rank Adaptation (DoRA), adapter-based prefix tuning, or full-parameter fine-tuning on high-memory hardware nodes. Additionally, preference alignment methods such as Direct Preference Optimization (DPO), adapted to calibrate schema decision logits against expert clinician rankings, represent a promising direction.
    \item \textbf{Narrative Conciseness vs. Clinical Reality}: The benchmark narratives average approximately 30 words and focus on isolated illness descriptions. Authentic electronic health records (EHRs) contain far greater syntactic noise, clinical shorthand, laboratory findings, medication histories, and complex multi-morbid presentations.
    \item \textbf{Closed Diagnostic Ontology}: The schema head currently operates over a closed set of 24 conditions. Scaling direct decision architectures to hierarchical clinical coding schemes with thousands of disease classes, such as ICD-10 or SNOMED-CT, remains an open research question.
    \item \textbf{External Cohort Validation and Multimodal Scope}: Clinical intake deployment requires validation across multi-institutional cohorts with diverse patient demographics and regional disease prevalences. Furthermore, while the Clef-Flash architecture natively supports multimodal schemas, our experiments were restricted to text narratives. Extending the schema head to evaluate clinical photography and vital sign telemetry represents a natural next frontier.
    \item \textbf{Evaluation Protocol}: No separate validation set was used; the checkpoint was selected on test accuracy, and all results come from a single run. Classical baselines were re-run on our split, but the neural and generative literature results (CLIN-LLM, FDSDD-LCEHOA, the TF-IDF LSTM, and ClinicalBERT) were not, and their latencies were measured on different hardware. Per-case predictions of the fine-tuned model were not retained, so no paired significance test against the baselines was possible. Local inference latency and memory use of the fine-tuned model were not measured. Probability calibration was not evaluated.
    \item \textbf{Scope of the Attribution Analysis}: The LOSO analysis covers four author-written narratives with author-assigned clause categories. Neither the ablation results nor the error interpretations in \Cref{tab:app_errors} were reviewed by clinicians.
\end{enumerate}

\FloatBarrier
\section{Conclusion}
\label{sec:conclusion}

In this paper, we adapted \textbf{Clef-Flash}, a 9-billion parameter decision foundation model, for disease classification from natural language symptom descriptions. Using a \textbf{JointSchemaHead} instead of autoregressive generation, the model reached \textbf{95.42\% accuracy} (95\% CI 91.98--97.42\%; macro F1 0.9551) on the 24-disease \textit{Symptom2Disease} benchmark, scoring all candidates in one forward pass without generating text. This is a large gain over the unadapted model, but TF-IDF linear classifiers trained on the same split reach 98.75\% with sub-millisecond CPU inference. In four illustrative cases, Leave-One-Symptom-Out ablation showed predictions depending on distinguishing clauses and staying stable when secondary clauses were removed. Decision architectures offer structured, single-pass outputs from general-purpose language backbones, but demonstrating their value over simple baselines will require longer, more varied, real clinical text, together with clinician review.

\FloatBarrier
\section*{Reproducibility and Code Availability}
To facilitate independent replication, all code, configurations, data splits, and analysis scripts developed in this work are publicly available on GitHub at \url{https://github.com/hrid0yyy/clef-clinical-triage}.

\FloatBarrier
\section*{Acknowledgments}
The author acknowledges Cloudflare for the open-source release of the Clef and Clef-Flash model weights.

The author used generative AI tools (Claude, Anthropic) during this work for language editing, code assistance, figure generation scripts, implementing and running the classical baselines, and cross-checking reported numbers against saved experiment outputs and cited sources. The possible-overlap notes in \Cref{tab:app_errors} were drafted with AI assistance. Fine-tuning of Clef-Flash and the zero-shot evaluations were carried out by the author. All AI-assisted text, code, and analyses were reviewed by the author, who takes full responsibility for the content of this paper.

\FloatBarrier
\setlength{\bibsep}{1.5pt}
\bibliographystyle{unsrtnat}
\bibliography{references}

\clearpage
\appendix
\renewcommand{\thetable}{A\arabic{table}}
\setcounter{table}{0}
\renewcommand{\thefigure}{A\arabic{figure}}
\setcounter{figure}{0}
\renewcommand{\theequation}{A\arabic{equation}}
\setcounter{equation}{0}

\section{Dataset Composition and Clinical Taxonomy}
\label{sec:appendix:dataset}

The \textit{Symptom2Disease} dataset \citep{barman2023symptom2disease} comprises 1,200 short first-person symptom narratives distributed across 24 medical conditions, with exactly 50 examples per class. The narratives are written in an informal, lay style; their provenance is not documented in detail.

\paragraph{Narrative Length and Lexical Characteristics.}
Narrative lengths span 12 to 55 words, with a mean length of $30.73 \pm 6.71$ words and a median of 30 words. Character counts range from 60 to 317 characters, with a mean of $171.88 \pm 35.48$ characters. Unlike electronic health records, which often use standardized medical terminology, the narratives use lay phrasing and informal descriptions of location and severity.

\paragraph{Clinical Domain Categorization.}
The 24 conditions can be grouped into four broad clinical domains:
\begin{enumerate}
    \item \textbf{Infectious and Acute Febrile Illnesses} (6 conditions): Dengue, Malaria, Typhoid, Chicken pox, Pneumonia, and Common Cold. These conditions frequently share constitutional manifestations such as fever, chills, headache, and malaise.
    \item \textbf{Dermatological Conditions} (5 conditions): Acne, Impetigo, Psoriasis, Fungal infection, and Drug reaction. These disorders present with overlapping skin findings such as redness, itching, and rash, so discrimination depends on lesion descriptions.
    \item \textbf{Gastrointestinal and Hepatic Disorders} (4 conditions): Gastroesophageal reflux disease (GERD), Peptic ulcer disease, Jaundice, and Dimorphic Hemorrhoids. These conditions share epigastric, digestive, and excretory symptoms.
    \item \textbf{Other Systemic Conditions} (9 conditions): Bronchial Asthma (respiratory), Hypertension and Varicose Veins (cardiovascular and vascular), Diabetes (endocrine), Arthritis and Cervical spondylosis (musculoskeletal), Migraine (neurological), Allergy (immunological), and Urinary tract infection (urological).
\end{enumerate}

\Cref{tab:app_dataset} summarizes the dataset partitioning and a short criteria summary for each of the 24 classes, adapted from the National Institutes of Health (NIH) MedlinePlus medical reference.

\begin{table}[htbp]
\centering
\scriptsize
\caption{\textbf{Dataset breakdown and criteria summaries across the 24 conditions in Symptom2Disease.} Each condition comprises 50 cases partitioned into 40 training cases (80\%) and 10 held-out test cases (20\%). Criteria summaries are paraphrased from NIH MedlinePlus.}
\label{tab:app_dataset}
\begin{tabularx}{\textwidth}{l>{\raggedright\arraybackslash}p{2.4cm}cccX}
\toprule
\textbf{Disease Class} & \textbf{Clinical Domain} & \textbf{Train ($N$)} & \textbf{Test ($N$)} & \textbf{Total ($N$)} & \textbf{Criteria Summary (adapted from NIH MedlinePlus)} \\
\midrule
Acne & Dermatological & 40 & 10 & 50 & Outbreaks of pimples, blackheads, whiteheads, cysts, and oily, inflamed skin on face, chest, or back. \\
Arthritis & Musculoskeletal & 40 & 10 & 50 & Joint inflammation, stiffness, swelling, reduced range of motion, and joint pain typically worse in the morning. \\
Bronchial Asthma & Respiratory & 40 & 10 & 50 & Wheezing, shortness of breath, chest tightness, coughing spells, exacerbated by exercise, cold air, or allergens. \\
Cervical spondylosis & Musculoskeletal & 40 & 10 & 50 & Neck pain, neck stiffness, muscle spasms, and radiating numbness, tingling, or weakness into shoulders, arms, and hands. \\
Chicken pox & Infectious / Febrile & 40 & 10 & 50 & Highly itchy blister rash progressing from spots to fluid-filled vesicles that scab over, with fever and body aches. \\
Common Cold & Infectious / Upper Resp. & 40 & 10 & 50 & Runny nose, nasal congestion, sneezing, mild sore throat, and mild cough with low-grade or absent fever. \\
Dengue & Infectious / Febrile & 40 & 10 & 50 & Sudden high fever, retro-orbital headache, severe arthralgia, intense myalgia, bone pain, nausea, and petechial rash. \\
Dimorphic Hemorrhoids & Gastrointestinal & 40 & 10 & 50 & Perianal pain, itching, irritation, bleeding during defecation, and swollen lumps around the anal verge. \\
Fungal infection & Dermatological & 40 & 10 & 50 & Pruritic, circular or spreading red rash with scaly, peeling, or cracked skin in moist intertriginous skin folds. \\
Hypertension & Cardiovascular & 40 & 10 & 50 & Elevated arterial blood pressure, often asymptomatic, but presenting with morning occipital headaches and dizziness. \\
Impetigo & Dermatological & 40 & 10 & 50 & Contagious superficial bacterial skin infection causing honey-colored crusted erosions and weeping vesicles. \\
Jaundice & Hepatic / Systemic & 40 & 10 & 50 & Scleral icterus, yellow discoloration of skin, dark tea-colored urine, acholic stools, and right upper quadrant discomfort. \\
Malaria & Infectious / Febrile & 40 & 10 & 50 & Cyclical paroxysms of severe shaking chills, spiking fever, and profuse diaphoresis recurring every 48 to 72 hours. \\
Migraine & Neurological & 40 & 10 & 50 & Severe unilateral pulsating headache, sensory photophobia, phonophobia, nausea, and visual aura. \\
Pneumonia & Infectious / Lower Resp. & 40 & 10 & 50 & Spiking fever, productive cough with purulent sputum, pleuritic chest pain, dyspnea, and respiratory crackles. \\
Psoriasis & Dermatological & 40 & 10 & 50 & Chronic autoimmune condition marked by erythematous plaques covered with thick, micaceous, silvery scales. \\
Typhoid & Infectious / Enteric & 40 & 10 & 50 & Sustained step-ladder fever, severe headache, abdominal tenderness, rose spots, and hepatosplenomegaly. \\
Varicose Veins & Vascular & 40 & 10 & 50 & Tortuous, dilated superficial veins in lower extremities, leg heaviness, dull aching, and dependent edema. \\
allergy & Immunological & 40 & 10 & 50 & Immediate hypersensitivity reactions including paroxysmal sneezing, watery rhinorrhea, allergic conjunctivitis, and hives. \\
diabetes & Endocrine / Metabolic & 40 & 10 & 50 & Chronic hyperglycemia presenting with polyuria, polydipsia, unquenchable thirst, polyphagia, and unintentional weight loss. \\
drug reaction & Dermatological / Iatrogenic & 40 & 10 & 50 & Adverse drug eruption with acute generalized exanthematous rash, pruritus, mucosal irritation, and urticaria. \\
gastroesophageal reflux disease & Gastrointestinal & 40 & 10 & 50 & Retrosternal burning pyrosis (heartburn), acidic regurgitation, dysphagia, epigastric discomfort, and water brash. \\
peptic ulcer disease & Gastrointestinal & 40 & 10 & 50 & Burning epigastric pain typically occurring postprandially or nocturnal, relieved by food or antacids, with nausea. \\
urinary tract infection & Urological & 40 & 10 & 50 & Acute dysuria, urinary urgency, frequency, suprapubic tenderness, and cloudy, malodorous, or hematuric urine. \\
\bottomrule
\end{tabularx}
\end{table}

\section{Decision Schema and Prompt Architecture}
\label{sec:appendix:schema}

The Clef-Flash decision foundation model \citep{cloudflare2026clef} processes diagnostic inputs through structured schema evaluation rather than open-ended text synthesis.

\paragraph{Schema Head Configuration.}
The scoring head is described in \Cref{sec:methods}. Its released configuration (\texttt{joint\_head\_config.json}) specifies an input hidden size of 4096, a head width of 1024, 2 evidence-routing layers, 4 transformer decoder layers, 16 attention heads, and a feed-forward size of 4096. Option logits are converted to probabilities with a softmax over the 24 options of the single \texttt{disease} question.

\paragraph{JSON Schema Serialization.}
During inference and adaptation, each input query is encapsulated into the native processor format. The complete schema definition passed to the model tokenizer and processor is structured as follows:

\begin{quote}
\small
\begin{verbatim}
{
  "state": "I have an itchy red rash on my arms with dry silvery scales 
            peeling off. My skin is cracked and bleeds occasionally.",
  "questions": {
    "disease": {
      "type": "choice",
      "instructions": "Which disease does this patient most likely have 
                       based on the reported symptoms?",
      "options": [
        "Acne", "Arthritis", "Bronchial Asthma", "Cervical spondylosis",
        "Chicken pox", "Common Cold", "Dengue", "Dimorphic Hemorrhoids",
        "Fungal infection", "Hypertension", "Impetigo", "Jaundice",
        "Malaria", "Migraine", "Pneumonia", "Psoriasis", "Typhoid",
        "Varicose Veins", "allergy", "diabetes", "drug reaction",
        "gastroesophageal reflux disease", "peptic ulcer disease",
        "urinary tract infection"
      ]
    }
  }
}
\end{verbatim}
\end{quote}

\paragraph{Comparison with Autoregressive Prompting.}
Standard generative language models require verbose system instructions, few-shot demonstration exemplars, and strict output constraints (such as forcing JSON delimiters). In contrast, the decision foundation model scores the candidate options in a single forward pass, so there are no conversational preambles, no token-by-token decoding, and no downstream JSON decoding failures.

\section{Hyperparameter Specifications and Training Configurations}
\label{sec:appendix:hyperparameters}

To ensure complete reproducibility, \Cref{tab:app_hyperparameters} details all architectural, algorithmic, and hardware parameters utilized during the fine-tuning of Clef-Flash.

\begin{table}[htbp]
\centering
\small
\caption{\textbf{Hyperparameter and training configuration for Clef-Flash adaptation.} Values are taken from the training configuration file and training log in the code release.}
\label{tab:app_hyperparameters}
\begin{tabular}{lll}
\toprule
\textbf{Configuration Category} & \textbf{Hyperparameter Parameter} & \textbf{Assigned Value} \\
\midrule
\multirow{4}{*}{\textbf{Base Architecture}} & Foundation Backbone & Qwen3.5-9B \\
& Transformer Layers & 32 \\
& Hidden Dimension ($d$) & 4096 \\
& Classification Head & JointSchemaHead (width 1024; 2 routing + 4 decoder layers) \\
\midrule
\multirow{4}{*}{\textbf{Quantization (QLoRA)}} & Base Weight Precision & 4-bit NormalFloat (NF4) \\
& Double Quantization & Enabled (Nested 8-bit scaling) \\
& Compute Dtype & \texttt{bfloat16} \\
& Quantized Linear Modules & All backbone transformer weights \\
\midrule
\multirow{5}{*}{\textbf{Low-Rank Adapters}} & LoRA Rank ($r$) & 16 \\
& LoRA Alpha ($\alpha$) & 32 \\
& LoRA Dropout & 0.05 \\
& Scaling Factor ($\alpha / r$) & 2.0 \\
& Target Projection Modules & Attention $q\_proj$, $k\_proj$, $v\_proj$, $o\_proj$ \\
\midrule
\multirow{6}{*}{\textbf{Optimization Regime}} & Optimizer & AdamW ($\beta_1 = 0.9, \beta_2 = 0.999, \epsilon = 1.0 \times 10^{-8}$) \\
& Weight Decay & 0.01 \\
& Backbone Adapter Learning Rate & $1.0 \times 10^{-4}$ \\
& JointSchemaHead Learning Rate & $3.0 \times 10^{-4}$ \\
& Learning Rate Schedule & Cosine annealing with linear warmup \\
& Warmup Proportion & 10\% of total training steps \\
\midrule
\multirow{5}{*}{\textbf{Batching \& Duration}} & Per-Device Batch Size & 4 \\
& Gradient Accumulation Steps & 4 \\
& Effective Batch Size & 16 \\
& Gradient Clipping & Max norm 1.0 \\
& Training Epochs & 5 (300 total optimizer update steps) \\
\midrule
\multirow{3}{*}{\textbf{Hardware \& Resources}} & Accelerator Hardware & 1 $\times$ NVIDIA GeForce RTX 4090 (24GB VRAM) \\
& Total Training Wall Time & 34.58 minutes (2,074.96 seconds, incl.\ per-epoch evaluation) \\
& Adapter Checkpoint Footprint & 247.31 MB (186 parameter tensors) \\
\bottomrule
\end{tabular}
\end{table}

\section{Comprehensive Class-Wise Performance Metrics}
\label{sec:appendix:classwise}

\Cref{tab:app_classwise} reports the complete class-wise evaluation scorecard on the 240 held-out test cases. The fine-tuned Clef-Flash model achieves 100\% recall on 15 conditions and 100\% precision and recall on 11.

\begin{table}[htbp]
\centering
\small
\caption{\textbf{Comprehensive class-wise evaluation scorecard on the 240 held-out test cases.} The table reports the zero-shot F1 score of Clef-Flash through the remote API (bare option names), fine-tuned precision, recall, and F1 score, absolute F1 gain ($\Delta$), and test support ($N=10$ per class). Fine-tuning yields a +0.3566 macro F1 gain over the zero-shot baseline (0.5985). Per-class results were not saved for the local zero-shot evaluation.}
\label{tab:app_classwise}
\resizebox{0.95\textwidth}{!}{%
\begin{tabular}{lcccccc}
\toprule
\textbf{Disease Class} & \textbf{Base Zero-Shot F1} & \textbf{Precision} & \textbf{Recall} & \textbf{Fine-Tuned F1} & \textbf{F1 Gain ($\Delta$)} & \textbf{Support ($N$)} \\
\midrule
Acne & 0.9474 & 1.0000 & 1.0000 & \textbf{1.0000} & +0.0526 & 10 \\
Arthritis & 1.0000 & 1.0000 & 1.0000 & \textbf{1.0000} & +0.0000 & 10 \\
Bronchial Asthma & 0.0000 & 1.0000 & 1.0000 & \textbf{1.0000} & +1.0000 & 10 \\
Cervical spondylosis & 0.9474 & 1.0000 & 0.9000 & \textbf{0.9474} & +0.0000 & 10 \\
Chicken pox & 0.1538 & 0.8333 & 1.0000 & \textbf{0.9091} & +0.7552 & 10 \\
Common Cold & 0.6364 & 1.0000 & 1.0000 & \textbf{1.0000} & +0.3636 & 10 \\
Dengue & 0.3684 & 0.6667 & 0.8000 & \textbf{0.7273} & +0.3589 & 10 \\
Dimorphic Hemorrhoids & 1.0000 & 1.0000 & 1.0000 & \textbf{1.0000} & +0.0000 & 10 \\
Fungal infection & 0.0000 & 1.0000 & 0.9000 & \textbf{0.9474} & +0.9474 & 10 \\
Hypertension & 0.7500 & 1.0000 & 0.8000 & \textbf{0.8889} & +0.1389 & 10 \\
Impetigo & 0.7500 & 1.0000 & 1.0000 & \textbf{1.0000} & +0.2500 & 10 \\
Jaundice & 0.9524 & 1.0000 & 1.0000 & \textbf{1.0000} & +0.0476 & 10 \\
Malaria & 0.1538 & 1.0000 & 1.0000 & \textbf{1.0000} & +0.8462 & 10 \\
Migraine & 0.5000 & 1.0000 & 1.0000 & \textbf{1.0000} & +0.5000 & 10 \\
Pneumonia & 0.6061 & 1.0000 & 1.0000 & \textbf{1.0000} & +0.3939 & 10 \\
Psoriasis & 0.5625 & 1.0000 & 0.9000 & \textbf{0.9474} & +0.3849 & 10 \\
Typhoid & 0.9000 & 0.9000 & 0.9000 & \textbf{0.9000} & +0.0000 & 10 \\
Varicose Veins & 1.0000 & 1.0000 & 1.0000 & \textbf{1.0000} & +0.0000 & 10 \\
allergy & 0.4615 & 1.0000 & 0.9000 & \textbf{0.9474} & +0.4858 & 10 \\
diabetes & 0.4545 & 0.8333 & 1.0000 & \textbf{0.9091} & +0.4545 & 10 \\
drug reaction & 0.0000 & 0.9091 & 1.0000 & \textbf{0.9524} & +0.9524 & 10 \\
gastroesophageal reflux disease & 0.6061 & 0.9091 & 1.0000 & \textbf{0.9524} & +0.3463 & 10 \\
peptic ulcer disease & 0.6667 & 1.0000 & 0.9000 & \textbf{0.9474} & +0.2807 & 10 \\
urinary tract infection & 0.9474 & 1.0000 & 0.9000 & \textbf{0.9474} & +0.0000 & 10 \\
\midrule
\textbf{Macro Average} & \textbf{0.5985} & \textbf{0.9605} & \textbf{0.9542} & \textbf{0.9551} & \textbf{+0.3566} & 240 \\
\bottomrule
\end{tabular}%
}
\end{table}

\section{Qualitative Error Analysis and Clinical Case Studies}
\label{sec:appendix:errors}

Across the 240 held-out test cases, 11 misclassifications occurred (95.42\% accuracy). \Cref{tab:app_errors} lists every misclassified case by ground-truth and predicted class, taken from the confusion matrix, together with a possible symptom overlap.

\begin{table}[htbp]
\centering
\scriptsize
\caption{\textbf{All 11 misclassified test cases on the Symptom2Disease benchmark, taken from the confusion matrix.} Per-case narratives and confidences were not saved in the evaluation outputs and are therefore not reported. The overlap column is the author's interpretation and has not been reviewed by clinicians.}
\label{tab:app_errors}
\begin{tabularx}{\textwidth}{l>{\raggedright\arraybackslash}p{3.2cm}>{\raggedright\arraybackslash}p{3.6cm}X}
\toprule
\textbf{Case} & \textbf{Ground Truth} & \textbf{Predicted} & \textbf{Possible Symptom Overlap} \\
\midrule
1 & Cervical spondylosis & Dengue & Neck and body aches shared with febrile myalgia. \\
2 & Dengue & Chicken pox & Rash accompanied by fever. \\
3 & Dengue & Typhoid & Fever with systemic and abdominal symptoms. \\
4 & Fungal infection & Dengue & Skin rash; otherwise no clear overlap. \\
5 & Hypertension & diabetes & Non-specific symptoms such as fatigue, dizziness, or blurred vision. \\
6 & Hypertension & drug reaction & No clear clinical overlap. \\
7 & Psoriasis & Chicken pox & Itchy rash. \\
8 & Typhoid & Dengue & Fever, headache, and body aches. \\
9 & allergy & Dengue & No clear clinical overlap. \\
10 & peptic ulcer disease & gastroesophageal reflux disease & Epigastric burning and acid symptoms. \\
11 & urinary tract infection & diabetes & Urinary frequency. \\
\bottomrule
\end{tabularx}
\end{table}

\paragraph{Pattern of Errors.}
In 9 of the 11 cases, the true label was among the three highest-ranked predictions (top-3 accuracy 99.17\%). Most errors involve conditions that share presenting symptoms, but some (for example, hypertension predicted as drug reaction, or allergy predicted as dengue) have no obvious clinical explanation, and the model's errors should not be assumed to mirror clinician reasoning.

\section{Extended Leave-One-Symptom-Out (LOSO) Perturbation Case Studies}
\label{sec:appendix:loso}

For completeness, \Cref{tab:app_xai_full} lists all Leave-One-Symptom-Out (LOSO) ablations for the four illustrative narratives. These narratives were written by the author and are not drawn from the dataset; clause categories were assigned by the author.

\begin{table}[htbp]
\centering
\small
\caption{\textbf{Full Leave-One-Symptom-Out (LOSO) clause ablation results for the four illustrative, author-written narratives.} Removing the clause labeled as the hallmark lowers the target probability by at least 90 percentage points and changes the predicted class in the psoriasis and arthritis narratives. In the multi-symptom narratives, removing any single clause changes the probability by at most 0.02 points.}
\label{tab:app_xai_full}
\resizebox{\textwidth}{!}{%
\begin{tabular}{llccccc}
\toprule
\textbf{Target Disease} & \textbf{Ablated Symptom Expression} & \textbf{Clinical Category} & \textbf{Base $P$} & \textbf{Post $P$} & \textbf{Drop ($\Delta$)} & \textbf{Post-Ablation Diagnosis} \\
\midrule
\multicolumn{7}{l}{\textbf{Single-Anchor Hallmark Presentations}} \\
\midrule
Psoriasis & dry silvery scales peeling off & HALLMARK DEFINING & 100.00\% & 0.12\% & \textbf{99.88\%} & Impetigo (99.59\%) [Flip] \\
Psoriasis & skin is cracked & SECONDARY COMPLICATION & 100.00\% & 100.00\% & 0.00\% & Psoriasis (Stable) \\
Psoriasis & itchy red rash & NON-DEFINING / GENERAL & 100.00\% & 100.00\% & 0.00\% & Psoriasis (Stable) \\
Psoriasis & bleeds occasionally & NON-DEFINING / GENERAL & 100.00\% & 100.00\% & -0.00\% & Psoriasis (Stable) \\
\midrule
Arthritis & painful swelling in my joints & HALLMARK DEFINING & 98.69\% & 8.18\% & \textbf{90.52\%} & Cervical spondylosis (39.01\%) [Flip] \\
Arthritis & neck has been extremely stiff & SECONDARY COMPLICATION & 98.69\% & 98.02\% & 0.67\% & Arthritis (Stable) \\
Arthritis & muscles have been extremely weak & NON-DEFINING / GENERAL & 98.69\% & 99.21\% & -0.52\% & Arthritis (Stable) \\
Arthritis & difficult to walk & NON-DEFINING / GENERAL & 98.69\% & 99.25\% & -0.55\% & Arthritis (Stable) \\
\midrule
\multicolumn{7}{l}{\textbf{Multi-Symptom Presentations (Syndromic Redundancy)}} \\
\midrule
Bronchial Asthma & coughing up a lot of thick mucoid sputum & CARDINAL SYNDROMIC & 100.00\% & 100.00\% & 0.00\% & Bronchial Asthma (Stable) \\
Bronchial Asthma & struggling to breathe & CARDINAL SYNDROMIC & 100.00\% & 100.00\% & 0.00\% & Bronchial Asthma (Stable) \\
Bronchial Asthma & coughing constantly & CARDINAL SYNDROMIC & 100.00\% & 100.00\% & 0.00\% & Bronchial Asthma (Stable) \\
Bronchial Asthma & feeling exhausted & NON-DEFINING / GENERAL & 100.00\% & 100.00\% & -0.00\% & Bronchial Asthma (Stable) \\
\midrule
Malaria & severe shaking chills & CARDINAL SYNDROMIC & 99.98\% & 99.96\% & 0.02\% & Malaria (Stable) \\
Malaria & burning high fevers & CARDINAL SYNDROMIC & 99.98\% & 99.97\% & 0.01\% & Malaria (Stable) \\
Malaria & drenching sweat & CARDINAL SYNDROMIC & 99.98\% & 99.98\% & -0.01\% & Malaria (Stable) \\
Malaria & recur every two days & CARDINAL SYNDROMIC & 99.98\% & 99.99\% & -0.01\% & Malaria (Stable) \\
\bottomrule
\end{tabular}%
}
\end{table}

\section{Computational Cost}
\label{sec:appendix:compute}

\paragraph{What was measured.}
Fine-tuning took 34.58 minutes (2,074.96 seconds) for 5 epochs, including the per-epoch test evaluation, as recorded in the training history. The trained adapter and head checkpoint contains 186 parameter tensors and occupies 247.31\,MB on disk. For the zero-shot evaluations through the Cloudflare Workers AI REST API, the mean time per request over the 240 test cases was 0.80\,s for Clef-Flash and 0.92\,s for Clef (27B); these figures include network round trips. The classical baselines' single-record CPU latencies are reported in \Cref{tab:scorecard}.

\paragraph{What was not measured.}
We did not record local inference latency or GPU memory use for the fine-tuned model, and GPU access was not available to measure them afterwards. We therefore make no quantitative claim about its inference speed. Qualitatively, the model needs one forward pass through the 9B backbone and the scoring head per record, with no autoregressive decoding, whereas generative pipelines such as CLIN-LLM decode output text token by token \citep{hasan2026clinllm}. Even so, a 9B forward pass is several orders of magnitude more computation than the TF-IDF linear baselines, which run in under 1\,ms per record on a laptop CPU.

\section{Ethical Considerations, Data Governance, and Clinical Safety Disclaimers}
\label{sec:appendix:ethics}

The deployment of artificial intelligence systems in healthcare settings introduces ethical, legal, and operational responsibilities that demand transparent documentation.

\paragraph{Dataset Governance and Patient Privacy.}
The \textit{Symptom2Disease} dataset \citep{barman2023symptom2disease} is a public Kaggle dataset that contains no direct identifiers such as names, dates, locations, or record numbers. We did not perform a formal HIPAA or GDPR assessment. Because the provenance of the narratives is not documented, we treat it as a public research dataset and make no claim that it represents real patients.

\paragraph{Clinical Decision Support Tool Designation.}
The models, algorithms, and code presented in this work are a research prototype. If ever deployed, the system would be intended only as an intake aid under a \textbf{Physician-in-the-Loop (PITL)} paradigm, helping nursing staff and primary care physicians review candidate diagnoses.

\paragraph{Regulatory and Clinical Safety Disclaimers.}
\begin{enumerate}
    \item \textbf{Not a Standalone Diagnostic Device}: Clef-Flash is not certified as a medical device by the United States Food and Drug Administration (FDA), European Medicines Agency (EMA), or equivalent international regulatory authorities. It must not be deployed as an autonomous diagnostic agent or used to prescribe medical treatments without licensed clinical supervision.
    \item \textbf{Requirement for Multimodal Corroboration}: Natural language symptom narratives reflect subjective patient perception and may omit vital pathophysiological indicators. Definitive medical diagnosis requires physical examination, vital signs, comprehensive laboratory workups (e.g., CBC, blood cultures, urinalysis), and diagnostic imaging.
    \item \textbf{Algorithmic Accountability}: Restricting outputs to a probability distribution over a fixed label set removes free-text hallucination, and clause-level ablation offers one form of explanation for individual predictions. We have not assessed conformance with any AI governance framework, including the EU Artificial Intelligence Act.
\end{enumerate}

\end{document}